\documentclass[runningheads]{llncs}
\usepackage{graphicx}
\usepackage{xcolor}
\usepackage{multirow}
\begin{document}
\title{Identifying Editor Roles in Argumentative Writing from Student Revision Histories}
\titlerunning{Identifying Editor Roles in Writing from Student Revision Histories}
%

\author{Tazin Afrin \and Diane Litman}

\authorrunning{T. Afrin and D. Litman.}
%

\institute{University of Pittsburgh, Pittsburgh, PA, USA 15260\\
\email{\{tazinafrin,litman\}@cs.pitt.edu}}
\maketitle              
\begin{abstract}
We present a method for identifying editor roles from students' revision behaviors during argumentative writing. 
We first develop a method for applying
a topic modeling algorithm to identify a set of editor roles from a vocabulary capturing three aspects of student revision behaviors: operation, purpose, and position. 
We validate the identified roles by showing that modeling the editor roles that students take when revising a paper not only accounts for the variance in revision purposes in our data, but also relates to writing improvement. 

\end{abstract}
\section{Introduction}
\label{introduction}
Knowing that experienced and successful writers revise differently than inexperienced writers~\cite{faigley1981w}, 
various intelligent writing  tools have been developed 
that provide localized feedback on text characteristics \cite{grammarly,elireview,roscoe2013m:writingpal,ets-writing-mentor}. 
These tools typically  suggest edits to guide revision, rather than model the editing process after observing revisions.
With the long term goal of developing an
intelligent revision assistant,
this paper presents an approach to modeling student editor roles.

Prior natural language processing (NLP) approaches to student revision analysis have focused on 
identifying revisions  during  argumentative writing and classifying their purposes and other properties ~\cite{zhang2015l,zhang2017hh,tan2014l,afrin2018improvement}. 
In contrast, editor roles have generally been studied in NLP using online collaborative writing applications such as 
  Wikipedia 
\cite{yang2016hkh}.  Inspired by the use of Wikipedia revision histories~\cite{yang2016hkh}, in this paper we similarly use topic modeling applied to  revision histories  to identify editor roles in the domain of student argumentative writing. 
To model student revision histories, between-draft essay revisions are extracted at a sentence-level and represented in terms of the following three aspects: operation (add, delete, or modify a sentence), purpose (e.g., correct grammar versus improve fluency), and position (revise at the beginning, middle or the end of an essay).  To identify  editor roles,  
a Latent Dirichlet Allocation (LDA)~\cite{blei2003nj} graphical model is then applied to these revision histories. 
Finally, we show that 
the identified roles capture the variability in our  data as well as 
correlate with writing improvement. 
\section{Corpora}
\label{corpus}
Our work takes advantage of several corpora of multiple drafts of argumentative essays written by both high-school and college students ~\cite{zhang2015l,zhang2017hh}, where all data
has been
annotated for revision using the framework of~\cite{zhang2015l}.
We divide our data into a Modeling Corpus (185 paired drafts, 3245 revisions) and an Evaluation Corpus (107 paired draft, 2045 revisions), based on whether expert grades are available before (Score1) and after (Score2) essay revision. 
Although the grading rubrics for the college and high-school essays in the Evaluation Corpus are different, both are based upon common criteria of argumentative writing, e.g., clear thesis, convincing evidence, clear wording without grammatical errors, 
etc. 
We apply linear scaling\footnote{Formula used to scale the scores = 100*(x-min)/(max-min).} to bring the scores within the same range of [0,100]. 
After scaling, the average Score1 and Score2 are 64.41 and 73.59, respectively. 

\begin{figure}[t]
\centering
\includegraphics[width=\textwidth]{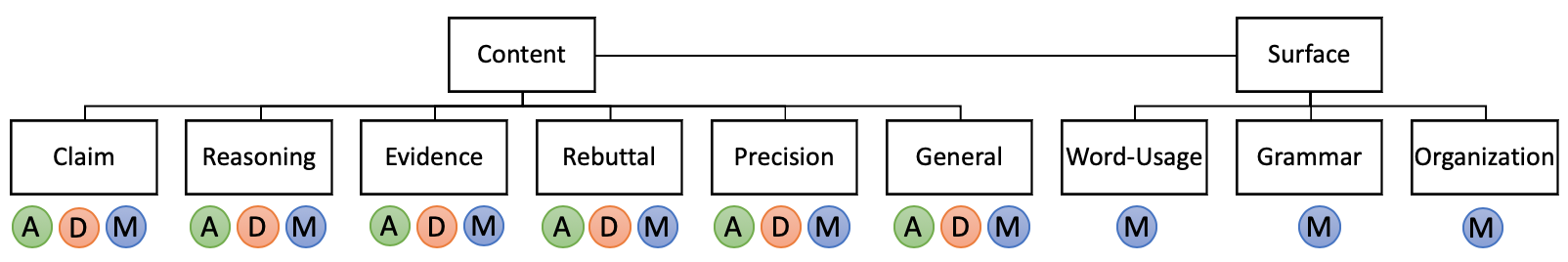}
\caption{The taxonomy of revision purposes~\cite{zhang2015l} (A:Add, D:Delete, M:Modify).} 
\label{fig:revision-schema}
\end{figure}

For all essays and prior to this study, subsequent drafts were manually aligned at the sentence-level based on semantic similarity.
Nonidentical aligned sentences 
were extracted as the {\bf revisions}, resulting in three types of revision {\bf operations} - $Add$, $Delete$, $Modify$. 
Each extracted revision was manually annotated with a {\bf purpose} following 
the revision schema shown in  Figure~\ref{fig:revision-schema} (modified compared to \cite{zhang2015l} by adding the Precision category). 
For this study, each revision's {\bf position} was in addition automatically tagged using its paragraph position in the revised essay. 
To maintain consistency across essays, instead of using paragraph number, we identify whether a revision is in the first ($beg$), last ($end$), or a middle ($mid$) paragraph. 
Table~\ref{table:purposes} shows a modified claim at the beginning of an essay from the Modeling Corpus.

\begin{table}[t]
\centering
\caption{Example revision from aligned drafts of an essay from the Modeling Corpus.}
\label{table:purposes}
\resizebox{\textwidth}{!}{%
\begin{tabular}{|p{4.5cm}|p{5.6cm}|l|l|l|}
\hline
 \textbf{Original Draft} &  \textbf{Revised Draft}      & \textbf{Operation} & \textbf{Purpose} & \textbf{Position}\\
\hline
 Self-driving vehicles pose many advantages and disadvantages. & While self-driving vehicles pose many advantages and disadvantages, I am not on the bandwagon for them at this time. & Modify & Claim & Beg. \\ \hline

\end{tabular}
}
\end{table}
\section{Identifying Editor Roles}
\label{modeling editor roles}
\begin{table}[t]
\centering
\caption{Derived editor roles  with top 10 revisions. (Blue: Surface, Orange: Content)}
\label{table:topics}
\resizebox{\textwidth}{!}{
\begin{tabular}{|c||c||c||c||c|}
\hline
\textbf{Proofreader}     & \textbf{Copy Editor}             & \textbf{Descriptive Editor}    & \textbf{Analytical Editor} & \textbf{Persuasive Editor} \\ \hline \hline
{\color{blue}Grammar\_mid}    & {\color{blue}Word-Usage\_mid}   & {\color{orange}+General\_mid}         & {\color{blue}Word-Usage\_beg}   & {\color{orange}+Reasoning\_mid}    \\ 
{\color{blue}Grammar\_beg}    & {\color{blue}Word-Usage\_beg}   & {\color{blue}Word-Usage\_mid}       & {\color{orange}+General\_end}     & {\color{orange}-Reasoning\_mid}    \\ 
{\color{blue}Word-Usage\_mid} & {\color{orange}+Reasoning\_mid}   & {\color{orange}-General\_mid}         & {\color{orange}+Reasoning\_end}   & {\color{orange}+Claims\_mid}      \\ 
{\color{blue}Grammar\_end}    & {\color{blue}Word-Usage\_end}   & {\color{orange}General\_mid}          & {\color{blue}Word-Usage\_end}   & {\color{orange}+Evidence\_mid}    \\ 
{\color{blue}Word-Usage\_end} & {\color{blue}Organization\_mid} & {\color{orange}Evidence\_mid}         & {\color{orange}Organization\_beg} & {\color{orange}+General\_mid}     \\ 
{\color{blue}Word-Usage\_beg} & {\color{orange}-General\_end}     & {\color{orange}Precision\_mid}        & {\color{orange}-Reasoning\_end}   & {\color{orange}-General\_mid}     \\ 
{\color{orange}Precision\_beg}  & {\color{orange}General\_end}      & {\color{orange}-General\_beg}         & {\color{orange}+Claims\_end}      & {\color{orange}Reasoning\_mid}    \\ 
{\color{orange}General\_mid}    & {\color{orange}-Reasoning\_mid}   & {\color{orange}+General\_beg}         & {\color{orange}+Evidence\_mid}    & {\color{orange}-General\_beg}     \\ 
{\color{orange}General\_end}    & {\color{orange}Claims\_mid}       & {\color{orange}Reasoning\_mid}        & {\color{orange}+Rebuttal\_end}    & {\color{orange}-Claims\_mid}      \\ 
{\color{orange}Reasoning\_beg}  & {\color{orange}-General\_mid}     & {\color{orange}+Claims\_beg}          & {\color{blue}Organization\_mid} & {\color{orange}+General\_beg}     \\ \hline
\end{tabular}
}
\end{table}
To create a vocabulary for topic modeling and to understand the repeating  patterns of student editors, we represent each revision utilizing the three aspects described earlier: operation, purpose, and position. 
This yields a rich and informative vocabulary for modeling our data, consisting of 63 revision ``words'' (54 content, 9 surface).  This is in contrast to the 24 word revision vocabulary used in the prior Wikipedia editor role extraction method~\cite{yang2016hkh}, 
formed using a Wiki-specific revision taxonomy of operation and purpose. 
When describing our revision ``words'', add and delete revisions are represented with `$+$' and `$-$' sign, and no sign for modification, e.g., $Claim\_beg$ in Table~\ref{table:purposes}. Editors are then represented by their history of revisions in terms of this revision  vocabulary. 

We trained the LDA model on the  Modeling Corpus and experimented with $2$ to $10$ topics.
After an extensive evaluation for topic interpretation based on top $10$ revisions under each topic, we ended up with $5$ topics where the revisions under each topic intuitively correspond to one of a set of potentially relevant editor roles for academic writing. 
We drew upon roles previously identified for writing domains such as newspaper editing  (e.g., proofreader, copy editor), Wikipedia (e.g., technical editor, substantive expert), and academic writing\footnote{https://sydney.edu.au/students/writing/types-of-academic-writing.html} (i.e., descriptive, analytical, persuasive, and critical).

The final topics are shown in Table~\ref{table:topics}, labeled by us with the best-matching editor role from the anticipated set of potential roles,  based on the  vocabulary items in each topic.
The defining characteristic of a {\bf Proofreader} are surface-level error corrections. 
{\bf Copy} editors ensure that the article is clear and concise as they revise for word-usage, clarity, and organization. 
{\bf Descriptive} editors  provide details and enhance clarity, with widespread  development of general content. 
{\bf Analytical} editors revise by adding information and better organizing thoughts, with top revision purposes being word-usage, content, reasoning, and rebuttal. 
{\bf Persuasive} editors discuss ideas and facts with relevant examples and develop arguments with added information. 
%
%
%
\section{Validating Editor Roles} 
\label{editor role validation}
Using the trained topic model, we first calculate the probability of an editor belonging to each of the 5 roles, for each editor in the Evaluation Corpus. These  probabilities represent each role's contribution to the essay revision. 
\begin{table}[t]
\centering
\caption{Variance across editors for each revision purpose (p$<.001:^{***}$, N=107).}
\label{table:topic validity regression}
\resizebox{\textwidth}{!}{%
\begin{tabular}{|l|c|c|c|c|c|c|c|c|}
\hline
\textbf{Purpose} & Grammar & Word-Usage &Organization & Claims	& Reasoning		& General	& Evidence	& Rebuttal\\ \hline
\textbf{$R^2$-value}	& $0.573^{***}$	& $0.537^{***}$	& $0.043$	& $0.240^{***}$	& $0.397^{***}$	& $0.459^{***}$	& $0.223^{***}$	& $0.025$ \\
\hline
\end{tabular}
}
\end{table}
Motivated by Wikipedia role validation~\cite{yang2016hkh}, we first validate our editor roles by similarly using editor roles to explain the  variance in revision purposes. 
We create 8 linear regression models, one for each revision purpose\footnote{The Evaluation Corpus does not have  precision revisions.}. The models take as input a five dimensional vector indicating an editor's contribution to each role and the output is the editor's edit frequency for each revision purpose. 
The R-squared values in Table~\ref{table:topic validity regression} show that our topic model can best explain the variance of Grammar, Word-Usage, General content, 
Claim, Reasoning, and Evidence edits. 

\begin{table}[t]
\centering
\caption{Partial correlations between role probabilities and Score2 controlling Score1.}
\label{table:par corr}
\resizebox{\textwidth}{!}{
\begin{tabular}{|c|c|c|c|c|c|}
\hline
\textbf{Editor Roles} 	& Proofreader & Copy & Descriptive & Analytical & Persuasive \\ \hline
\textbf{Corr(p-value)}	&	-0.175(0.073) & -0.049(0.621) & -0.180(0.064) & -0.013(0.891) & 0.205(\textbf{0.035}) \\
\hline
\end{tabular}
}
\end{table}

A corpus study in~\cite{zhang2015l} showed that content changes are correlated with argumentative writing improvement, reaffirming the statement of~\cite{faigley1981w}. 
Using a similar method, we investigate if our editor roles are related to writing improvement. 
We calculate partial Pearson correlations between editor roles and Score2 while controlling for Score1 to regress out the effect of the correlation between Score1 and Score2 (Corr.= $0.692$, $p < 0.001$). Table~\ref{table:par corr} shows that the roles consisting of only surface edits or a mixture of edits are not correlated to writing improvement. 
However, Persuasive editor, which consists of content revisions, shows a positive significant correlation to writing improvement. Our results suggest that the Persuasive editor is the role of an experienced writer.

%
%
%
\section{Conclusion and Future Work}

Although editor roles have been studied for online collaborative writing~\cite{yang2016hkh,welser2011ckl}, our research investigates  student revisions of  argumentative essays. 
While our model follows  previous methods~\cite{yang2016hkh}, we introduce a unique vocabulary to model each editor's revision history, 
with evaluation results suggesting that our identified roles  capture salient features of writing. 
Future plans include 
using a Markov model to consider revision order, expanding the revision vocabulary, 
and using the predictions to provide feedback in an intelligent revision assistant.

\subsection*{Acknowledgements}
This work is funded by NSF Award 1735752.

%
%
%
\bibliographystyle{splncs04}
\bibliography{references}

\end{document}